\documentclass[journal,10pt]{IEEEtran} % use the `journal` option for ITherm conference style
\IEEEoverridecommandlockouts
\usepackage{amsmath,amssymb,amsfonts,amsthm}
\usepackage{algorithm,algorithmicx}
\usepackage[noend]{algpseudocode} 
\usepackage{graphicx}
\usepackage{epstopdf}
\usepackage{xcolor}
\usepackage{cite}
\usepackage[left=1.62cm,right=1.62cm,top=1.78cm,bottom=2.33cm]{geometry}
\usepackage{latexsym,todonotes}
\usepackage{graphicx}
\usepackage{epstopdf}
\usepackage{float}
\usepackage{multirow}
\usepackage{txfonts}
\usepackage{mathptmx}
\usepackage{mathrsfs}
\usepackage{exscale}%
\usepackage{relsize}%

\usepackage{soul}

\usepackage{tikz,pgfplots}
\usepackage{flowchart}
\usepackage{tabularx}%
\usepackage{booktabs}%
\usepackage[switch]{lineno}%
\usepackage{bbding}%
\usepackage[bookmarks=false]{hyperref}
\definecolor{lavendergray}{rgb}{0.77, 0.76, 0.82}
\def\BibTeX{{\rm B\kern-.05em{\sc i\kern-.025em b}\kern-.08em
    T\kern-.1667em\lower.7ex\hbox{E}\kern-.125emX}}

\begin{document}

\title{Modeling and Joint Optimization of Security, Latency, and Computational Cost in Blockchain-based Healthcare Systems
%\thanks{Identify applicable funding agency here. If none, delete this.}
}

\author{%%%% author names
    \IEEEauthorblockN{Zukai Li\IEEEauthorrefmark{1}}% first author
    , \IEEEauthorblockN{Wei Tian\IEEEauthorrefmark{1}}% delete this line if not needed
    , \IEEEauthorblockN{Jingjin Wu\IEEEauthorrefmark{1}\IEEEauthorrefmark{2}}% delete this line if not needed
    % duplicate the line above as many times as needed to list all authors
    \\%%%% author affiliations
    \IEEEauthorblockA{\IEEEauthorrefmark{1}Department of Statistics and Data Science, BNU-HKBU United International College, Guangdong, P. R. China}\\% first affiliation
    \IEEEauthorblockA{\IEEEauthorrefmark{2}Guangdong Provincial Key Laboratory of Interdisciplinary Research and Application for Data
Science, Guangdong, P. R. China.}\\% delete this line if not needed
    % duplicate the line above as many times as needed to list all affiliations
    %%%% corresponding author contact details
    \IEEEauthorblockA{Email: p930005029@mail.uic.edu.cn, s230202702@mail.uic.edu.cn, jj.wu@ieee.org}
}

\maketitle

\thispagestyle{empty}
 
\begin{abstract}
    %The integration of Internet of Things (IoT) and blockchain technology can potentially improve the efficiency of healthcare management to a significant extent. 
    In the era of the Internet of Things (IoT), blockchain is a promising technology for improving the efficiency of healthcare systems, as it enables secure storage, management, and sharing of real-time health data collected by the IoT devices. As the implementations of blockchain-based healthcare systems usually involve multiple conflicting metrics, it is essential to balance them according to the requirements of specific scenarios. In this paper, we formulate a joint optimization model with three metrics, namely latency, security, and computational cost, that are particularly important for IoT-enabled healthcare. However, it is computationally intractable to identify the exact optimal solution of this problem for practical sized systems. Thus, we propose an algorithm called the Adaptive Discrete Particle Swarm Algorithm (ADPSA) to obtain near-optimal solutions in a low-complexity manner. With its roots in the classical Particle Swarm Optimization (PSO) algorithm, our proposed ADPSA can effectively manage the numerous binary and integer variables in the formulation. We demonstrate by extensive numerical experiments that the ADPSA consistently outperforms existing benchmark approaches, including the original PSO, exhaustive search and Simulated Annealing, in a wide range of scenarios. 
    
    %A large volume of data is generated in the operation of healthcare systems, and the judicious sharing of this data can be of significant benefit to all participants in the system. In order to ensure the security and traceability of healthcare data sharing, this paper proposes a blockchain-based structure for handling healthcare-related data. This structure allows data to be exchanged securely within the healthcare system, creating a positive cycle in the exchange of relevant healthcare data and thus making full use of data resources to create value. In particular, we have constructed an optimisation model based on the operational mechanism of our proposed \textcolor{red}{XXX structure} from the perspective of the system manager, and a new mixed algorithm, ADPSA is designed for the variable characteristics of the model. The solved optimal solution guides managers to flexibly allocate the available resources to achieve the goal of making the most efficient and appropriate processing of different types of data, enabling them to meet the needs of system entities in different scenarios. The result shows, our proposed algorithm ADPSA has \textcolor{red}{ .....}
    %Finally, we summarise the benefits of the \textcolor{red}{ XXX} processing structure and highlight the efficiency of the ADPSA algorithm for solving specific types of optimal problems and possible directions for future research.
\end{abstract}


\vspace{-0.15cm}
\section{Introduction}

%\timothy{IoT}

The Internet of Things (IoT) has played a crucial role in the recent development of healthcare management~\cite{selvaraj2020challenges}. IoT devices, such as wearable fitness trackers and connected medical devices, are capable of collecting and transmitting vast amounts of real-time health data, such that the customers can receive more personalized and efficient care from healthcare providers. Additionally, IoT technology can be used to automate various processes in healthcare management, such as supply chain management, patient monitoring, and appointment scheduling, to increase efficiency compared to conventional approaches.

More recently, IoT has been integrated with blockchain technology to further optimize the entire healtcare management process. Blockchain is a decentralized technology used for recording and verifying real-time transactions~\cite{agbo2019blockchain}. With a high level of privacy protection, it was originally developed for  facilitating cryptocurrency transactions, but was later also identified highly compatible for validation, storage and sharing of sensitive healthcare data collected by IoT devices~\cite{kuo2017blockchain}. So far, blockchain technology has been adopted in IoT-enabled healthcare systems for a number of applications, including automated claim authentication~\cite{angraal2017blockchain} and public health management~\cite{mettler2016blockchain}. %Through this system, different stakeholders are able to share health data seamlessly and securely, which enhancing the efficiency of data transmission and decision-making. Hence, numerous proposals for the application of blockchain in healthcare systems have been made, as the methods required to manage data related to healthcare are highly compatible with the functions of blockchain technology.  

The actual implementation of a blockchain-based system requires a meticulous evaluation of multiple potentially conflicting factors. For example, adding more verifiers would enhance the security of the verification process. However, it would incur extra computational cost and may prolong the verification time if one or more newly added verifiers are slower than the existing ones. Meanwhile, the weights of different factors may alter across different applications. For example, telemedicine~\cite{albahri2021iot} requires real-time interaction and thus would consider low latency as the most desirable attribute, while the security management of electronic health records~\cite{ray2021biothr} would accord a higher priority to security and privacy preservation as it encompasses a vast amount of sensitive data. As a blockchain manager, it is essential to determine the optimal blockchain configuration for a certain application scenario. 

%the verification and configurations are natural for blockchain managers to allocate the available resources effectively to function efficiently and appropriately within the constraints. As the critical factors in blockchain-based healthcare systems, such as security, latency, and computational cost, often conflict with each other, it is necessary to formulate a joint optimization model to identify trade-offs among the factors and find the most suitable configuration for particular combinations of different weights to these factors.
%By referring to the optimal solution obtained by solving the optimisation model, the manager can decide how to properly allocate the available resources and let the system choose the most appropriate way to process different data types. 

%\zaaak{A little bit weird to put it here, coherence problem.}

Another common approach to improve the performance of IoT-based healthcare systems is to adopt hardware components with higher computational capabilities. For example, the Artificially Intelligent Electronic Money introduced in \cite{fragkos2020artificially} can support security assurance with a low computational complexity, by utilizing secure hardware in an offline scenario and engaging multiple trusted third parties to execute the binding process. However, hardware-based solutions generally incur additional replacement and deployment costs, and are therefore considered more expensive than the approach that we will focus on in this paper, which configures on existing system architecture. %as well as capital costs due to the involvement of trusted third parties and the manufacturing expenses of the hardware. 
%In addition, blockchain-based system architectures are more extensible than hardware solutions. Therefore, in this paper we focus on blockchain-based solutions.

In this paper, we investigate the practical operation of a blockchain-based healthcare system under the DPoS consensus scheme~\cite{abdellatif2020sshealth}, and formulate a joint optimization model with consideration to security, latency, and computational cost. The model is based on the actual operating mechanisms of blockchain-based healthcare systems enabled by IoT. From the perspective of a blockchain manager, we particularly focus on the impact caused by the selection of verifiers who would participate in validating the blockchain, and the size of a single block. The relative importance of the three metrics can be flexibly adjusted for different scenarios, by changing the weights in the formulation.

The complexity of solving the joint optimization problem increases exponentially with the number of candidate verifiers. Therefore, the exhaustive search method that was used in existing studies (e.g.~\cite{abdellatif2020sshealth}) to solve simpler formulations are deemed unsuitable. In light of this, we propose an intelligent algorithm, Adaptive Discrete Particle Swarm Algorithm (ADPSA), to obtain quasi-optimal solutions that balance the three metrics in a low-complexity manner.  While ADPSA has its roots in the traditional Particle Swarm Optimization (PSO)~\cite{kennedy1995particle}, we add special mechanisms that can effectively manage the numerous binary and integer variables in the problem. We demonstrate that ADPSA consistently outperforms the original PSO, the exhaustive search method, and Simulated Annealing (SA)~\cite{kirkpatrick1983optimization} even when the weights of the three metrics vary substantially.

The rest of this paper is organized as follows. Section II explains the formulation of the joint optimization model in detail. Section III describes the procedures and rationales of our proposed ADPSA. Experimental setup and numerical results are illustrated in Section IV. Finally, Section V concludes the paper.

%—————————————————————————————————————————————————————————————————————————————————————————————————————————————————————————————————————————————————%

\vspace{-0.15cm}
\section{Joint Optimization Model of Blockchain-based Healthcare Systems}

\vspace{-0.3cm}
\begin{figure}[htbp]
    \centering
    \includegraphics[scale=0.14]{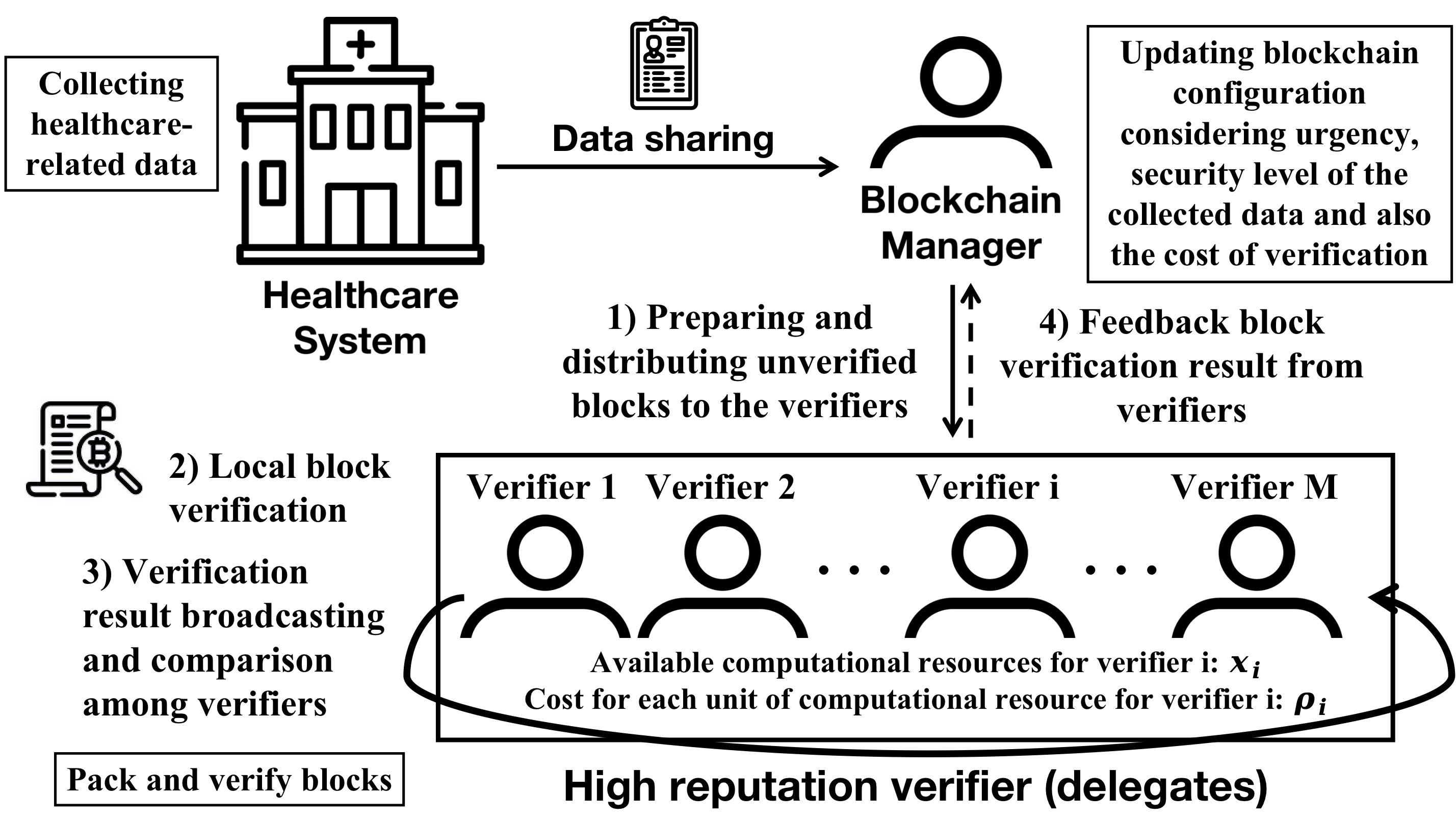}
    \caption{Blockchain-based processing of healthcare data under the DPoS consensus scheme.}
    \label{DPoS}
\end{figure}

On the basis of the DPoS consensus mechanism and the blockchain application framework described in \cite{abdellatif2020sshealth} and \cite{kang2019toward}, we summarize the main procedures for processing healthcare-related data on blockchain in Fig.~\ref{DPoS}. %The blockchain manager can choose different processing strategies for different types of data, to take advantage of the blockchain technology in managing and using medical data more efficiently. 

Our focus in this paper is to find the optimal blockchain configuration to efficiently process and transmit health data under different application scenarios. We consider the following three important factors:

%Therefore, it is necessary to formulate a joint optimization model to help blockchain managers to accurately measure the trade-off among the important factors that conflict each other, and find the most suitable configuration for particular combinations of different weights to these factors.

%\textcolor{red}{Is this description OK?}
\begin{itemize}
\item Latency ($L$): The total amount of confirmation time required for a block to be successfully packed;
\item Security ($S$): A quantified measurement that evaluates the risk of the blockchain being breached during the packing process. A higher value of $S$ indicates that the process is more secure;
\item Computational cost ($C$): Total monetary cost incurred to the blockchain manager for the entire packing process.
\end{itemize}

We now elaborate each of the factors in details.

%—————————————————————————————————————————————————————————————————————————————————————————————————————————————————————————————————————————————————%

%In the following, we will define latency, security and cost, respectively.

\vspace{-0.1cm}
\subsection{Latency}

Latency refers to the total time spent on the four main steps of the blockchain validation process: (1) Downlink transmission time for passing initial unvalidated blockchains from the blockchain manager to the verifiers; (2) Verification time for the slowest verifier to complete the verification process locally; (3) Validation time for the fastest verifier to broadcast the results to other verifiers, and other verifiers to cross-validate the result; (4) Uplink transmission time for the verifiers to pass the feedback on verification to the blockchain manager. 

Overall, the latency (in seconds) is defined by the following formula,

\begin{equation}
    L=\frac{\theta \cdot R}{v_{d}}+\max_{i \in\{1, \cdots, M\}}\left(\frac{K}{z_{i} \cdot x_{i}}\right)+\phi \cdot \theta \cdot R \cdot m+\frac{P}{v_{u}},
    \label{L}
\end{equation}

\noindent
where the four addend terms correspond to the four main steps described in Fig.~\ref{DPoS}.
In~\eqref{L}, $v_d$ is the downlink transmission rate from the block manager to the verifiers and $v_u$ is the uplink transmission rate from the verifiers to the block manager. $R$ is the size of one transaction, $\theta$ is the number of transactions in each block, and therefore $\theta \cdot R$ represents the size of the block. %The transmission time of unvalidated blockchain from the blockchain manager to the verifiers is $\frac{\theta \cdot R}{v_{d}}$. 
$K$ is the total amount of computational resources required to support the function of block verification, $x_i$ is the number of such resources that are available for verifier $i$, and $z_i$ is a binary variable that denotes whether the $i$th verifier is selected into the blockchain to perform verification and validation. Since the total time for verification is determined by the slowest one to finish the task, the total time for all verifiers to complete the verification is $\max\left(\frac{K}{z_{i} \cdot x_{i}}\right)$. %\zaaak{max is right here.} \timothy{revise.} 
According to \cite{kang2019toward} and~\cite{abdellatif2020sshealth}, the required time to broadcast and validate results among all verifiers depends on the size of the blockchain, the number of verifiers, and the speed of verification. Here, $\phi$ is an empirically determined constant and can be treated as an estimation of the average verification time of all verifiers~\cite{kang2019toward}. Finally, $P$ is the data size of verification feedback, and the transmission time of feedback on verification from the verifiers to the blockchain manager is then $\frac{P}{v_{u}}$.

\vspace{-0.1cm}
\subsection{Security}

We consider a generalizable model that quantifies the security. We define the security rating (dimensionless) as,

\begin{equation}
    S=\alpha \cdot m^{\kappa}.
    \label{S}
\end{equation}

The security of the system is considered to be positively correlated to the number of participating verifiers $m$. By reasonably assuming that most verifiers are faithful and the computational powers of all verifiers are similar in blockchain-based healthcare systems, we can infer that adding more verifiers would lower the risk that malicious verifiers occupy the majority of computational power and thus manipulate the whole system \cite{wust2016security}. In~\eqref{S}, $\alpha$ is a parameter based on the factors related to specific system implementations, such as the chosen blockchain type and the consensus protocol used, and $\kappa$ is a factor indicating the scale of the system \cite{kang2019toward}.

\vspace{-0.1cm}
\subsection{Computational cost}

We define the computational cost (in dollars) as

\begin{equation}
    C=\frac{\sum_{i=1}^{M} \rho_{i} \cdot z_i \cdot x_i}{\theta},
    \label{C}
\end{equation}

\noindent
where $\rho_{i} \cdot z_i \cdot x_i$ is the $i$th verifier's computational cost. $\rho_i$ is a coefficient that determines the cost paid by the $i$th verifier for each unit of computational resources (e.g., the cost for buying or renting computing servers, electricity fee). Such cost will be eventually borne by the blockchain manager, who needs to provide the incentive to the verifiers for them to participate in the process.

\vspace{-0.1cm}
\subsection{Joint optimization model}

With the above three factors, we formulate the joint optimization problem as a utility maximization problem:

\begin{equation}
    U=\beta_1 \cdot \frac{L_m-L}{L_{m}}+\beta_2 \cdot \frac{S}{S_m}+\beta_3 \cdot \frac{C_m-C}{C_{m}},
    \label{U}
\end{equation}

\noindent
where $\beta_{1}$, $\beta_{2}$ and $\beta_{3}$ are weights indicating the relative importance of the three metrics, with the normalizing condition that $\beta_{1}+\beta_{2}+\beta_{3}=1$. The specific values of the weights can be adjusted based on the preference of the blockchain manager or the customer requirements. Also, as the three metrics have different units, they are further normalized according to their respective maximum possible values $L_{m}$, $S_{m}$ and $C_{m}$, such that they are in a common scale in the utility function.

It is straightforward to observe that, by the normalization process and restrictions on the weights, the  range for $U$ is between $0$ and $1$. In this sense, the objective function value represents a linear combination of the relative ``scores" for the three normalized metrics. For a feasible solution, the closer the utility function value is to $1$, the more desirable that the system operates for the blockchain manager and/or customers.

%\textcolor{orange}{Would it be more natural for the efficiency function to be as large as possible?}

With the description of the model components above, our joint optimization model is formulated as follows,

\begin{equation}
    \begin{array}{ll}        \underset{m,\theta,\mathbf{z}}{\text{Maximize}} & U \\
        \text {s.t.} & m \in \{m_{\text{min}}, \ m_{\text{min}}+1, \ \cdots, \ m_{\text{max}}-1, \ m_{\text{max}}\}, \\
        & \theta \in \{\theta_{\text{min}}, \ \theta_{\text{min}}+1, \ \cdots, \ \theta_{\text{max}}-1, \ \theta_{\text{max}}\}, \\
        & z_i \in \{0, \ 1\} \quad {\text{for}} \quad i \ \in \{m_{\text{min}}, \cdots, m_{\text{max}}\}, \\
        & \sum_{i=1}^{m_{\text{max}}}z_i = m %\quad {\text{for}} \quad i \ \in \{1, \cdots, m_{\text{max}}\}.
    \end{array}
    \label{O_M}
\end{equation}

In our joint optimization model, there are two integer decision variables, $m$ and $\theta$, as well as $m_{\text{max}}-m_{\text{min}}+1$ binary decision variables, $\mathbf{z} = \{z_1, z_2, \cdots, z_{m_{\text{max}}-1}, z_{m_{\text{max}}\}}$. Recall that, $m$ is the number of verifiers selected, whose possible values are the integers from $m_{\text{min}}$ to $m_{\text{max}}$, while $\theta$ is the number of transactions in each block with minimum value $\theta_{\text{min}}$ and maximum value $\theta_{\text{max}}$, respectively. It is worth noting that all three components (latency, safety, cost) in \eqref{O_M} are normalized and therefore, dimensionless.

Compared with existing similar models such as those in  \cite{abdellatif2020sshealth} and \cite{kang2019toward}, our model innovatively includes the binary variables $\mathbf{z}$ in the calculation of the latency, in order to make the model more practical and interpretable. Intuitively, only verifiers who are selected to participate in a certain round of verification would have an impact on the overall latency, and the speed of the slowest selected verifier would dictate the verification time as reflected in the second term of~\eqref{L}. Meanwhile, by adjusting the weights $\beta_1, \beta_2$, and $\beta_3$, our formulation can be applied to a wide range of scenarios when the relative importance of the three metrics varies across different data characteristics or regions.

%Blockchain managers can adjust the functionality and utility of the blockchain by changing the values of $m$ and $\theta$.

%In comparison to the model mentioned in \cite{Abdellatif} and (add [12]), we innovatively introduce $M-v+1$ 0-1 variables $z_i$ in order to enhance the rationality and interpretability of the model. The blockchain manager can choose the most suitable verifiers from the available resources to enter the blockchain system with the help of the results of the 0-1 variables in the optimal solution, instead of selecting randomly or in a specific order.

%—————————————————————————————————————————————————————————————————————————————————————————————————————————————————————————————————————————————————%

\vspace{-0.15cm}
\section{Algorithm for the Optimal Blockchain Configuration}

The above formulation in~\eqref{O_M} is more applicable to real-world situations than existing models in previous research by taking more factors into consideration. However, the inclusion of the additional decision variables, especially the binary variables $\mathbf{z}$ also increases the complexity to solve the problem, which is a non-linear integer program by nature. Meanwhile, the number of decision variables can be very large when there exist many candidate verifiers (large $m_{\text{max}}$), making it almost impossible to find the optimal solution by exhaustive search. %\st{A brute-force exhaustive approach will have to check the feasibility of $O(2^{m_{\text{max}}}\theta_{\text{max}})$ possible solutions, and determine the optimality by computing the utility value for each feasible solution. Such operations would be very computationally intensive for systems of practical sizes.}

Note that the algorithm introduced in~\cite{abdellatif2020sshealth} to solve a reduced version of~\eqref{O_M} is not suitable for solving our problem. That algorithm is in a similar manner to exhaustive search and thus cannot address the complexity issue and is not guaranteed to converge to the global optimum. On the other hand, due to its discrete nature, the objective utility function~\eqref{U} cannot be differentiated, and its concavity cannot be determined by conventional methods. Therefore, traditional optimization algorithms such as gradient descent and Newton’s method are not applicable in this case either.

%\st{Therefore, this paper uses intelligent optimisation algorithms to address the above challenges.

%In terms of the principles on which they are based, intelligent optimisation algorithms can be divided into three main categories: evolutionary based, physics based and Swarm Intelligence based. From each of the three categories, we selecte the most basic genetic algorithms: genetic algorithms, annealing algorithms and particle swarm algorithms. 
%}

%\vspace{-0.5cm}
\vspace{-0.1cm}
\subsection{Overview of the proposed ADPSA algorithm}

Due to the introduction of the binary variable $\mathbf{z}$, changes in the binary variable during the search process can further cause large fluctuations in the value of the objective function. In addition, compared to continuous variables, it is much more difficult to decide the direction and rate of change during the iterative process for binary variables. These problems lead to the increase in both the complexity and instability of the algorithm. 

Therefore, to address the above challenges, we focus on the Particle Swarm Optimization (PSO) algorithm, which is particularly suitable for problems with a large number of decision variables and easy to modify according to the requirements. Since the original PSO algorithm was mainly used to solve optimization problems with continuous variables, we made further improvements to the algorithm. Based on the traditional PSO algorithm, we devise new update formulas for velocities and positions of particles, to make the proposed algorithm applicable to discrete variables. We refer to the proposed algorithm as Adaptive Discrete Particle Swarm Algorithm (ADPSA). The pseudo-code of ADPSA is presented in Algorithm~\ref{al:ADPSA-algo}. Some key notations are explained as follows.

%\textcolor{red}{Check the pseudocode again}

%\zaaak{The $x$ in the model and the position $x$ in the algorithm are reused.}

\renewcommand{\algorithmicrequire}{\textbf{Input:}}  % Use Input in the format of Algorithm
\renewcommand{\algorithmicensure}{\textbf{Output:}} % Use Output in the format of Algorithm

\begin{algorithm}[h!]

  \begin{algorithmic}[1]
    \Require
        $c_1, c_2,n,N$
        %$v_d, v_u, R, K, x, \phi, P, \alpha, \kappa, \rho, n, N, m_{\text{max}}, m_{\text{min}}, \theta_{\text{max}}, \theta_{\text{min}}$. %\timothy{some of the inputs seem not used in the algorithm.}
    \Ensure
        $g^*$, fit$(g^*)$.
      
    \State Randomly initialize  $\mathbf{v}^0_j$ and $X^0_j$ for all $j$.

    \State Initialize  $p_j^{*} = X_j$ for all $j$, and set $g^{*} = \underset{j}{\max}\{p^{*}_j\}$.
    %\State Initialize the optimal population solution $\mathbf{g}^{*} =$ max\{$p^{*}$\}

    % \While {$T > T_{\text{min}}$ and \textcolor{red}{flag == 1}}
    
    \For{$i = 1$ to $n$}

        %\State the updated velocity of discrete variables $m$ and $\theta$
        \State Generate $r_j$ and $s_j$ from Uniform$(0,1)$ for all $j$.
        \For{$j=1$ to $N$}
            
        \State $v^i_{j,m} = w_i  v^{i-1}_{j,m} + c_1 r_{j} (p^*_{j,m} - m^{i-1}_j) + c_2  s_{j} (g^{*}_m - m^{i-1}_j)$;

        \State $v^i_{j,\theta} = w_i  v^{i-1}_{j,\theta} + c_1 r_{j} (p^*_{j,\theta} - \theta^{i-1}_j) + c_2  s_{j} (g^{*}_\theta - \theta^{i-1}_j)$;
        
        \State $m^i_j =[m^{i-1}_j + v^i_{j,m}]$;

        \State $\theta^i_j =[\theta^{i-1}_j + v^i_{j,\theta}]$;

        %\State \textcolor{red}{update binary variables $z^i_j$ according to subsection $D$ in this section}

        \EndFor

        % \State $\mathbf{x_{tempij}}$ = $\mathbf{h\(x_{ijz^-}\)}$;

        % \If {fit($\mathbf{x_{tempij}}$ $<$ fit($\mathbf{x_{(i-1)j}}$)}
            
        %     \State $\mathbf{x}$ = $\mathbf{x_{tempij}}$
                
        % \Else
            
        %     \If {random[0,1]  $<$ selection probability $P$}
                
        %         \State $\mathbf{x_{ij}} = \mathbf{x_{tempij}}$;
                    
        %     \Else
                
        %         \State $\mathbf{x_{ij} = x_{(i-1)j}}$
                    
        %     \EndIf
        % \EndIf
        \State Set $z^i_{jk} = 0$ for $k = 1, 2, \cdots, M$;
        \State Set $N_m = m^*$;

        \State Sample $N_m$ elements without replacement from $\{z^i_{j1}, z^i_{j2}, \cdots, z^i_{jM}\}$, and set their values to $1$;

        \If {fit$(X^i_j)$ $>$ fit$(p^*_j)$}
            
            \State Update $p^*_j = X^i_j$;
                
        \EndIf
            
        \If {fit($p^*_j$) $>$ fit($g^{*}$)}
            
            \State Update $g^{*} = p^*_j$;
                
        \EndIf
    \EndFor
    
% \EndWhile

   % \State \zaaak{output($\mathbf{g}^{*}$,  fit$(\mathbf{g}^{*})$)}
    
  \end{algorithmic} 
  \caption{ADPSA}   
\label{al:ADPSA-algo}
\end{algorithm}

%In our optimization model, the variables are all non-negative integers, where all other variables are 0-1 variables except for the variables $m$ and $\theta$. In order to improve the efficiency and convergence accuracy of the algorithm, it becomes critical to update the variables reasonably in each iteration of the search process. 

\begin{itemize}

\item $p^{*}_{j}$: the optimal position ever reached by the $j$th particle since the beginning of the searching process;

\item $X^{i}_{j}$: the position for the $j$th particle at the $i$th iteration, in the form of $X^i_j = (m^i_j, \theta^i_j, z^i_{j1}, z^i_{j2}, \dots, z^i_{j(m_{\text{max}}-1)}, z^i_{jm_{\text{max}}})$;

\item $\mathbf{v}^{i}_{j}$: the velocity for the $j$th particle at the $i$th iteration, in the form of $\mathbf{v}^{i}_{j} = (v^i_{j,m}, v^i_{j,\theta}, v^i_{j,z_1}, v^i_{j,z_2}, \dots, v^i_{j,z_{m_{\text{max}}-1}}, v^i_{j,z_{m_{\text{max}}}})$;

\item $g^*$: the optimal position solution obtained by the model, in the form of $g^* = (m^{*}, \theta^{*}, z_{1}^*, z_{2}^*, \dots, z_{m_{\text{max}}-1}^*, z_{m_{\text{max}}}^*)$;

\item fit($X^i_j$): the objective function value achieved by a feasible solution $X^i_j$;

\item $[x]$: the nearest integer of a real number $x$;

\item $N$: the population size of the particle swarm.

\end{itemize}

\vspace{-0.1cm}
\subsection{Generation of the initial particle swarm}

We now highlight several key aspects of ADPSA that help to overcome the difficulties when applying the original PSO in solving~\eqref{O_M}. 

Before the iterative search starts, we initialize the position and velocity of the particle swarm in the following way.

\vspace{-0.1cm}
\subsubsection{Position}

We generate the initial population in a more uniform way in the feasible solution space. Recall that, the position of a particle represents a set of feasible solutions in the form $X = (m, \theta, z_{1}, z_{2}, \dots, z_{M-1}, z_{M})$. For the first two position coordinates $m$ and $\theta$, we initialize them in the following way:

\begin{equation}
    \Omega_{pm} = \left\{\left[  \frac{i \cdot (m_{\text{max}}-m_{\text{min}})}{c_{pm}} \right]\right\} \quad \text{for} \ i \ \in \{1, \cdots, N\},
    \label{pm_ini}
\end{equation}

\begin{equation}
    \Omega_{p\theta} = \left\{\left[  \frac{i \cdot (\theta_{\text{max}}-\theta_{\text{min}})}{c_{p\theta}}  \right] \right\}\quad \text{for} \ i \ \in \{1, \cdots, N\},
    \label{ptheta_ini}
\end{equation}

\noindent
where $c_{pm}$ and $c_{p\theta}$ are constants that can be adjusted according to the population size $N$, while $m_{\text{max}},m_{\text{min}},\theta_{\text{max}}$ and $\theta_{\text{max}}$ represent the maximum and minimum allowable $m$ and $\theta$ values, respectively. We enumerate all possible binary combinations of the elements in $\Omega_{pm}$ and $\Omega_{p\theta}$, and assign each combination as the initial values $m^0_j$ and $\theta^0_j$ for particle $j$.

Then, for particle $j$, we randomly set $m^0_j$ binary variables of $z^0_j$ to $1$, and the rest of the binary variables would be assigned an initial value of $0$. This setting allows the particle swarm to cover the feasible solution space in a more uniform way as compared to randomly generated initial positions, reducing the risk of missing the optimal solution, especially when a smaller $N$ is applied under the computational power constraint.

%\zaaak{Whether setting the initial position in this way can optimize the algorithm is yet to be verified.}

\subsubsection{Velocity}

The velocity represents the search direction of a particle at a particular iteration. For particle $j$, the velocity is denoted as $(v^{j}_{m}, v^{j}_{\theta}, v^{j}_{z_1}, v^{j}_{z_2}, \dots, v^{j}_{z_{m_{\text{max}}-1}}, v^{j}_{z_{m_{\text{max}}}})$. Due to the specificity of the ADPSA algorithm (which will be explained in Subsection D), we only need to generate the initial velocity for searching $m$ and $\theta$ by

\begin{equation}
    v^0_{j,m} = r_{m}  \frac{m_{\text{max}}-m_{\text{min}}}{c_{vm}},
    \label{vm_ini}
\end{equation}

\begin{equation}
    v^0_{j,\theta} = r_{\theta}  \frac{\theta_{\text{max}}-\theta_{\text{min}}}{c_{v\theta}},
    \label{vtheta_ini}
\end{equation}

\noindent
where $r_{m}$ and $r_{\theta}$ are random numbers generated from the Uniform $[0,1]$ distribution, $c_{vm}$ and $c_{v\theta}$ are constants that can be adjusted according to the range of variable restrictions. The generation of the initial velocity by~\eqref{vm_ini} and~\eqref{vtheta_ini} ensures the rationality of the searching step size as well as the randomness of the search, as a suitable step size is essential for the algorithm to search finely and efficiently without getting trapped in a local optimum solution.

\subsection{Update strategies for $m$ and $\theta$}

One key modification of our proposed APDSA from the original PSO is that, we revise the formula for updating the velocity at each iteration for the variables $m$ and $\theta$. Specifically, for the $j$th particle at iteration $i$, its velocity and position with respect to variables $m$ and $\theta$ are updated by Lines 6 -- 9 in Algorithm~\ref{al:ADPSA-algo}. Note that, as the constraint for $z_i$ depends on the value of $m$, the update of $z_i$ values would occur after $m$ is determined in each iteration.
% \begin{equation}
%     v^{i}_{j}=w \cdot v^{i-1}_j + c_{1} \cdot r_{j} \cdot (\mathbf{p}^{*}_{j}-\mathbf{x}^{i-1}_{j}) + c_{2} \cdot r_{2} \cdot (\mathbf{g}^*-\mathbf{x}^{i-1}_{j}),
%     \label{velocity_pso}
% \end{equation}

% \begin{equation}
%     \mathbf{x}^{i}_{j}=\mathbf{x}^{i-1}_{j} + \mathbf{v}^{i}_{j}
%     \label{position_pso}
% \end{equation}

% \noindent
% % where $\mathbf{x}^{i-1}_{j}$ is the position of the $j$th particle at the beginning of iteration $i$.

% \timothy{Original PSO cannot handle integer variables - use rounding.}

According to the original PSO, $w^i_{j,m}$, $c_1$, $c_2$ are constant terms that impact the weights of the velocity of the $j$th particle at the previous iteration, $v^{i-1}_j$, the best position in the searched history of the same particle, $p^{*}_{j}$, and the best position that had been reached by any particles, $g^{*}$,  respectively. Meanwhile, $r_j$ and $s_j$ are two stochastic terms that introduce randomness to the searching process, aiming to reduce the probability that the algorithm stops at a local optimum. Here, we let both $r_j$ and $s_j$ follow the uniform distribution $(0,1)$.

In APDSA, we apply a slightly different update approach, that is, we consider that the inertia terms $w^i_{j,m}$ are also updated at each iteration. At iteration $i$, we have 

\begin{equation}
    w_i = w_{\text{max}} - \frac{i(w_{\text{max}}-w_{\text{min}})}{n}.
    \label{w}
\end{equation}

Compared to the original PSO, we modify the inertia weight to the form presented in~\eqref{w}. Here $w_{\text{max}}$ and $w_{\text{min}}$ are the maximum and minimum values of the inertia term; we set them to $0.9$ and $0.4$ in this paper, respectively, according to the recommendation of \cite{yasuda2003adaptive}. Updating the inertia term by~\eqref{w} will make it gradually decrease as the search proceeds. In this way, we allow the algorithm to cover a wide range of solutions and avoid falling into a local optimum at the beginning of the search, and converge to an optimal solution faster after more iterations have been processed. Also, to ensure that the updated variables are integers, a rounding mechanism is added as in Lines 8 and 9 in Algorithm~\ref{al:ADPSA-algo}, which is considered as a common approach when integer variables are involved in PSO~\cite{rezaee2015particle}. %\timothy{add ref for rounding.}. %We also take the lower or upper bound of the variables when they are outside the limit after updating.

\vspace{-0.1cm}
\subsection{Update strategies for binary decision variables}

As discussed in the previous subsection, in solving our joint optimization model~\eqref{O_M}, the searching processes for the discrete variables $m$ and $\theta$ can follow a similar manner as in the original PSO algorithm. However, it is not appropriate to apply the same mechanism for binary decision variables in our problem because the sum of all binary variables is restricted to $m$.

% An intuitive explanation is that, if we handle $z_i$ in the same way as $m$ or $\theta$, $z_i$ would be randomly assigned a value of $0$ or $1$ in each iteration, which does not add any useful information to approach the optimal solution to the previous iterations. 

Consider the binary variable of the $j$th particle at the $i$th iteration are $z^{i-1}_j$ = [$z_{j1}^{i-1}, z_{j2}^{i-1}, \dots, z_{j(m_{\text{max}}-1)}^{i-1}, z_{jm_{\text{max}}}^{i-1}$]. There are exactly $m$ variables which would be assigned the value of $1$ among $z_1$ to $z_{m_{\text{max}}}$. After updating the velocity and the position of $m$ and $\theta$ in the same iteration, to guarantee the efficiency and accuracy of the searching process, we propose to process all these binary variables with the following steps (corresponding to Lines $10$ to $12$ in Algorithm~\ref{al:ADPSA-algo}).

\noindent
For the $j$th particle of the $i$th iteration:

\begin{enumerate}
    \item Determine the number of binary variables $N_m$ to be randomly generated based on the updated $m$ from the previous iteration.

    \item Randomly generate $N_m$ distinct integers from $\{1,2,\cdots,m_{\text{max}}\}$. Denote the selected set as $\Omega_L$.

    \item Set all $z_k$ to 1 for $k$ $\in$ $\Omega_L$ and  obtain the updated binary variable $z^i_j$ = $(z_{j1}^i, z_{j2}^i, \dots, z_{j(m_{\text{max}}-1)}^i, z_{jm_{\text{max}}}^i)$.
\end{enumerate}

After particle $j$'s binary variable position is updated as above, we combine all the updated positions of variables to get the final position $X^i_j = (m^i_j, \theta^i_j, z^i_{j1}, z^1_{j2}, \cdots, z^i_{j(m_{\text{max}}-1)}, z^i_{jm_{\text{max}}})$ of the $j$th particle in iteration $i$.

\vspace{-0.1cm}
\subsection{Complexity analysis}

%\zaaak{the section title?}

%\zaaak{complexcity analysis: describe and compare PSO and ADPSA}

From Algorithm~\ref{al:ADPSA-algo}, we can infer that the computational complexity of the ADPSA is $O(n)$, which is the same as the original PSO. An intuitive explanation for the complexity is that, compared to the original PSO, our major modifications in ADPSA include a new method of initializing the population, revised updating rules for the particle positions during each iteration based on the features of decision variables, and enhanced searching process by adaptive inertia weights. The skeleton of PSO algorithm, however, remains unchanged in the ADPSA. Comparatively, a brute-force exhaustive search approach will have to check the feasibility and optimalty of all possible combinations, leading to a computational complexity of $O(2^{m_{\text{max}}}\theta_{\text{max}})$.

%—————————————————————————————————————————————————————————————————————————————————————————————————————————————————————————————————————————————————%

\vspace{-0.15cm}
\section{Numerical Results}

\begin{figure}[htbp]
    \centering
    \includegraphics[scale=0.21]{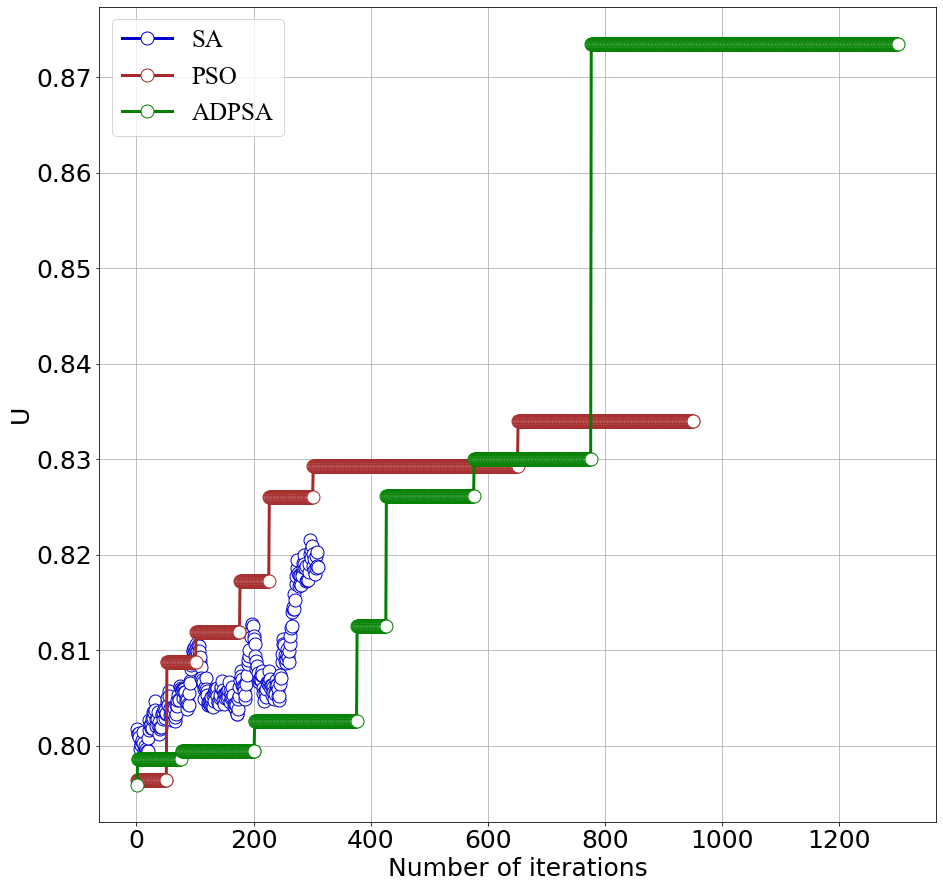}
    \caption{A demonstration of objective function value of SA, PSO, and ADPSA with respect to the number of iterations.}
    \label{fig:result_1r}
\end{figure}

In this section, we evaluate the performance of ADPSA algorithm numerically. We first conduct a one-round test using a fixed set of simulation parameters shown in Table \ref{table:1} compared the performance of the three algorithms: SA, PSO and the ADPSA. We fix the total time of every iteration to $1$ second, and the results are shown in Fig.~\ref{fig:result_1r}. Based on the results, it can be seen that ADPSA performs more iterations than the other two basic intelligence algorithms within the same time limit, and achieves the highest utility value.

\vspace{-0.25cm}
\begin{table}[h!]
    \begin{center}
    \caption{Simulation parameters}
    \label{table:1}
        \begin{tabular}{|c|c|c|c|}
            \hline Parameter & Value & Parameter & Value \\
            \hline $m_{\text{min}}$  & 2 &  $m_{\text{max}}$  & 1000 \\
            \hline $\theta_{\text{min}}$  & 2 &  $\theta_{\text{max}}$  & 1000 \\
            \hline $v_{d}$  &  1.2 $\mathrm{Mb}$ / $\mathrm{s}$  &  $v_{u}$  &  1.3 $\mathrm{Mb}$ / $\mathrm{s}$  \\
            \hline $R$  &  2 $\mathrm{kb}$  &  $P$  &  0.5 $\mathrm{Mb}$  \\
            \hline K  &  59292098.2754  &  $\phi$  &  0.5  \\
            \hline $\alpha$  &  5  &  $\kappa$  &  1  \\
            \hline $\rho$ & \multicolumn{3}{|c|}{(99.9318, 92.6047, ...,  100.9061) \$/$\mathrm{kb}$}  \\
            \hline $X$ & \multicolumn{3}{|c|}{(34329.04315, 44717.13606, ..., 42341.5001) $\mathrm{kb}$}  \\
            \hline ($\beta_1,\beta_2,\beta_3$) & \multicolumn{3}{|c|}{(4/10, 2/10, 4/10)}  \\
            \hline
        \end{tabular}
    \end{center}
\end{table}

\vspace{-0.8cm}

% Next, we conduct a total of 100 rounds of trials using the parameters in Table \ref{table:1}, comparing the four algorithms of pseudo-exhaustive method, SA, PSO and ADPSA, and Figure \ref{result_100r} shows the results of each of the four methods. As to the pseudo-exhaustive method, we first disregard all the binary variables and use the exhaustive method to select the optimal combination of variables $m$ and $\theta$, and then randomly generate the binary variables that meet the restrictions according to the value of $m$. The results of the pseudo-exhaustive method demonstrate that the discrete variables $m$ and $\theta$ affect the objective function values simultaneously with all binary variables, so they cannot be considered separately or selected partially for optimization first. Although the SA algorithm search the best results of the four algorithms, it also obtains poor results in many rounds and performs extremely unstable. We believe its performance is highly dependent on the initial position and speed. In comparison, the ADPSA algorithm has the best overall performance and steadily searches for superior solutions, with larger mean, quartiles and smaller variance than other methods.

\begin{figure}[h!]
    \centering
    \includegraphics[scale=0.21]{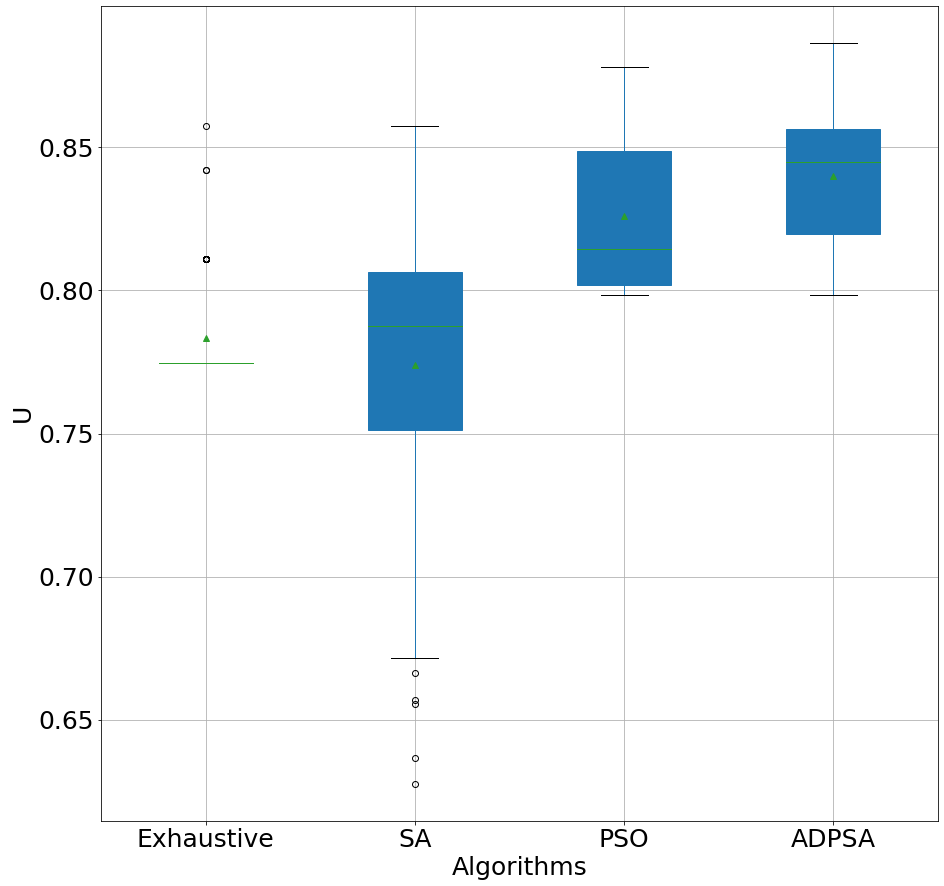}
    \caption{Box plot of experiment results with 100 trials from each algorithm.}
    \label{result_100r}
\end{figure}

\vspace{-0.15cm}
Next, we compare a pseudo-exhaustive search method, SA, PSO and ADPSA, and Fig. \ref{result_100r} shows the results of each of the four methods by conducting a total of 100 rounds of trials using the parameters in Table \ref{table:1}. As to the pseudo-exhaustive method, we first disregard all the binary variables and use the exhaustive method to select the optimal combination of variables $m$ and $\theta$, and then randomly generate the binary variables that meet the restrictions according to the value of $m$. The results of the pseudo-exhaustive method demonstrate that the discrete variables $m$ and $\theta$ affect the objective function values simultaneously with all binary variables, so they cannot be considered separately or selected partially for optimization first. %Although the SA algorithm search the best results of the four algorithms, it also obtains poor results in many rounds and 
For SA, the performance is extremely unstable and highly dependent on the initial position and speed. In comparison, the ADPSA algorithm achieves the best overall performance, with larger mean, quartiles and smaller variance than other methods.

\begin{figure}[h!]
    \centering
    \includegraphics[scale=0.21]{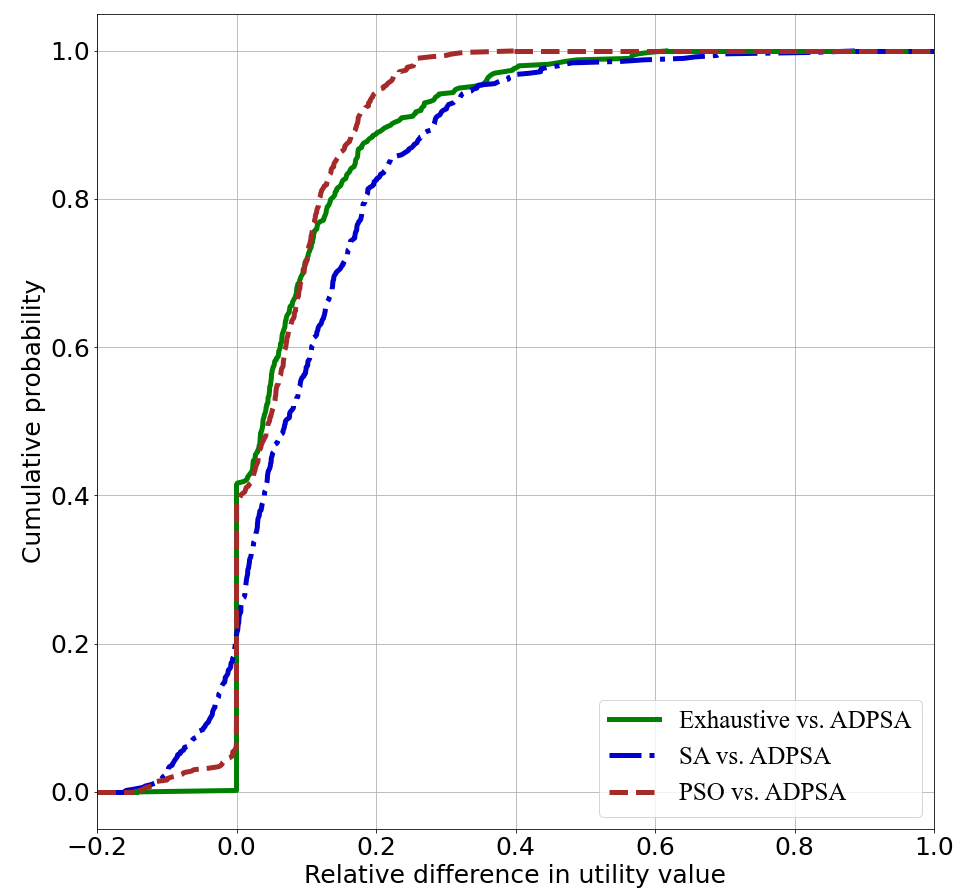}
    \caption{Cumulative probability distribution of the relative difference in the utility value by ADPSA to other algorithms with randomly generated weights.}
    \label{diffbeta_result}
\end{figure}

Finally, we test the performance of each algorithm when the blockchain manager assigns different weights to Latency, Security, and Cost according to different requirements. %The parameters used in the test are the same as in Table \ref{table:1}, except that $\beta_1$, $\beta_2$ and $\beta_3$ are adjusted. We randomly generate 500 different sets of weights and conduct a trail on each set of weights. The final results are shown in Figure~\ref{diffbeta_result}. We can observe that the ADPSA algorithm still performs well for different combinations of weights, \zaaak{further explain how significant this difference in performance between the two algorithms is}.
Fig.~\ref{diffbeta_result} shows the cumulative probability distribution of the relative difference of ADPSA to other algorithms, defined as $\frac{U_\text{ADPSA}-U_A}{U_A}$, where $A$ can be one of PSO, SA and exhaustive search, for $500$ independent runs with randomly generated weights $\beta_1, \beta_2$, and $\beta_3$. We observe that ADPSA performs the same or better than PSO, SA, and exhaustive search in 
81.2\%, 78.8\%, 99.8\% of all scenarios. The improvement in the utility value of ADPSA over PSO, SA, and exhaustive search can be up to 39.7\%, 88.7\%, and 61.6\%, respectively.

% \begin{table}[h!]
%     \begin{center}
%     \caption{Weight combinations of 10 test cases}
%     \label{table:2}
%         \begin{tabular}{|c|c|c|c|}
%             \hline Combination number & $\beta_1$ & $\beta_1$ & $\beta_1$ \\
%             \hline 1 & 0.57 & 0.23 & 0.20 \\
%             \hline 2 & 0.43 & 0.24 & 0.33 \\
%             \hline 3 & 0.50 & 0.20 & 0.30 \\
%             \hline 4 & 0.26 & 0.20 & 0.54 \\
%             \hline 5 & 0.34 & 0.46 & 0.20 \\
%             \hline 6 & 0.44 & 0.48 & 0.08 \\
%             \hline 7 & 0.10 & 0.47 & 0.43 \\
%             \hline 8 & 0.40 & 0.36 & 0.24 \\
%             \hline 9 & 0.25 & 0.28 & 0.47 \\
%             \hline 10 & 0.50 & 0.41 & 0.09 \\
%             \hline
%         \end{tabular}
%     \end{center}
% \end{table}

\vspace{-0.3cm}

% \vspace{4ex}
% \begin{table}[h!]
%     \begin{center}
%     \caption{Results of 100 rounds of trials}
%         \begin{tabular}{|c|c|c|c|}

%             \hline  & PSO & SA & PSAA \\
%             \hline Mean & 0.1863 & 0.2351 & 0.2167 \\
%             \hline Variance & 0.001978 & 0.002437 & 0.002276 \\
%             \hline Number of iterations (in 1s) & 3952 & 1981 & 87219 \\
%             \hline Time spent for one iteration & 0.002529 & 0.005046 & 0.000114 \\
%             \hline
            
%         \end{tabular}
%     \end{center}
%     \label{results}
% \end{table}

\vspace{-0.2cm}
The improvement of ADPSA over the original PSO algorithm is mainly due to the modified initial population generation and particle property update methods, which were described in detail in subsections B, C and D in Section III. The improved algorithm can search for the optimal solution to the problem more comprehensively and precisely.

%—————————————————————————————————————————————————————————————————————————————————————————————————————————————————————————————————————————————————%

\vspace{-0.15cm}
\section{Conclusions}

In this paper, we have introduced a method for processing healthcare-related data using blockchain technology based on the DPoS consensus algorithm, ensuring a balance of security and efficiency needed for healthcare systems. To help blockchain managers better handle and manage data, we have designed an optimization model based on the blockchain operational process, and proposed the ADPSA algorithm based on the PSO algorithm to address the presence of discrete variables and a large number of binary variables in the model. Numerical results have shown that ADPSA is superior as well as more efficient and robust than existing algorithms.

% With our designed optimization model and the ADPSA algorithm, blockchain managers can select the most appropriate blockchain configuration by the type of data and the optimal solutions of the model to achieve the goal of making the system work efficiently. As a result of the systems and models we have constructed, healthcare data will be leveraged in a more secure and efficient manner, providing greater value to participants in the healthcare system. 
%In the future, we can look for more accurate and efficient algorithms to improve the performance of handling a large number of discrete variables as well as binary variables in combinatorial optimization problems.

%\textcolor{red}{Is it mainly a case of recapitulating the content of the article or focusing on what contributions have been made on the basis of the references?}

%\textcolor{red}{Future Directions refer to:}

%—————————————————————————————————————————————————————————————————————————————————————————————————————————————————————————————————————————————————%

%\section*{Acknowledgment}

%The preferred spelling of the word ``acknowledgment'' in America is without
%an ``e'' after the ``g''. Avoid the stilted expression ``one of us (R. B.
%G.) thanks $\ldots$''. Instead, try ``R. B. G. thanks$\ldots$''. Put sponsor
%acknowledgments in the unnumbered footnote on the first page.

\section*{Acknowledgements}

This work is partly supported by Zhuhai Basic and Applied Basic Research Foundation Grant ZH22017003200018PWC, and partly supported by the Guangdong Provincial Key Laboratory of 
Interdisciplinary Research and Application for Data Science, BNU-HKBU United International 
College, Project code 2022B1212010006 and in part by Guangdong Higher Education Upgrading
Plan (2021-2025) UIC  R0400001-22.

\bibliographystyle{IEEEtran}
\bibliography{ref}

\vspace{12pt}

\end{document}